# XEns-CKD: An Explainable Ensemble-Based Approach for Chronic Kidney Disease Stage Detection

Rehan Ahmad†[1], Gousia Habib†*, Muhammad Shaban[3], and Ishfaq Ahmad Malik[4]

[1] SVKM's, NMIMS, Mukesh Patel School of Technology Management and Engineering, Shirpur, Maharashtra, India.
[*2]FCAI Fellow University of Helsinki and ELLIS Institute Helsinki, Finland,
[3] COER University, Roorkee, India,
[4]Shoolini University Solan Himachal Pradesh, India
rehan.ahmad@nmims.edu , gousia.habib@helsinki.fi , mshaban0121@gmail.com, malikishfaq7@gmail.com

*Corresponding Author: Gousia Habib**

**†Equal Contribution**

***Abstract:*** *Chronic kidney disease (CKD) is a silent disease. Its progression does not hamper a person's daily routine. A human kidney function can be classified in normal or in one of the five stages of CKD. Early detection of CKD stage can help patient to know the functional status of the kidney and follow medical advice to slowdown the CKD progression. In this paper we propose a novel XEns-CKD (Ensemble vision transformer based) scheme for CKD stage classification from ultrasound images. Three ViT's trained on private ultrasound CKD image dataset. Each ViT tuned on different training parameters. The performance of each ViT estimated using macro sensitivity, macro specificity, macro precision, macro F-1 score, macro Youden Index, MCC and macro balanced accuracy. The ensemble model results demonstrate that overall classification accuracy is 86.36%. The emphasis of this work is to identify and interpret the affected kidney areas because of CKD progression. For result interpretation task, explainable artificial intelligence (XAI) techniques like LIME, LRP, Attention_Min., Attention_Max. used to enhance model transparency and clinical trust. An attention map combining Attention_Min., Attention_Max. results identify and interpret strongly the affected areas of kidney because of CKD progression from one stage to another. Attention map also highlights the impact of CKD progression in these areas. When the results of proposed method compared with the existing methods it is observed that, proposed method identifies CKD stages and Normal state of kidney with an increment of 4% in accuracy.*

**Keyword:** CKD, Ultrasound Kidney Image, ViT, Ensemble, XAI, GRAD-CAM, LRP, LIME, SHAP

## 1 Introduction

Chronic kidney disease (CKD) is a silently progressing disease [1]. The progression of CKD not felt by the person. When realized, CKD had reached to a complicated state. CKD destroys the functioning capability of human kidney. CKD can be identified using a clinical test known as glomerular filtration rate test (GFR test) [2]. GFR test processes patient's blood and urine sample to identify the presence of toxicity in human urine filtered by kidney. The functioning state of human kidney can be identified based on GFR value in ml/min/1.73 $m^2$ units. A human kidney is declared healthy if its GFR value is in between 100 ~ 120. GFR values in the range of 90 ~ 99, 60 ~ 89, 30 ~ 59, 15 ~ 29 and <15 represent CKD stage-1, stage-2, stage-3 stage-4 and stage-5, respectively [3]. Stage-1 is an indication of primary stage of CKD. Stage-2 indicate the kidney disfunction started. Stage-3 indicates the critical stage of kidney function started. Stage-4 indicates kidney has fallen to a critical level. And stage-5 indicate kidney is no longer functional to support the body [4].

GFR test has several shortfalls like, it is expensive, invasive, causing discomfort to patient and requires time to produce result, therefore mostly avoided [5]. An image-based technique is cheap, non-invasive and produces result quickly. CKD stage identification from ultrasound (US) does not cause any discomfort to patient and is a preferable choice by the clinicians [6]. The issue of despeckling in US image can be taken care by use of appropriate filter [3].

In the last decade medical image processing domain witnessed tremendous growth in acquisition, management and storage of images. Artificial intelligence (AI) algorithms capable to process thousands of images simultaneously for clinical decision. AI assisted systems aids clinicians in decision support system [7]. The optimization and advancement in AI algorithms improved the accuracy of AI assisted systems. AI systems works like a black box. From their behavior it is not clear how it reaches to a decision [8]. AI models can be improved in terms of transparency and interpretability and the rationale for their produced output [9]. In medical domain, the decision of AI algorithms is under scrutiny due to black box behavior, and an explanation about the achieved result will help clinicians to reach on decision and ease of interpretation. Use of explainable artificial intelligence (XAI) is a solution [10]. XAI is a subdomain of AI, which offers transparency, interpretability and explainability for the results produced [9]. Several works have been proposed by the researchers in past few years for use of XAI for CKD detection. They all used different method of explanation of results (decision). All these methods of XAI are discussed here to identify the key issue in this domain.

A scheme for diagnosis of kidney diseases (normal, cyst, tumor, stone) EfficientNetV2, InceptionNetV2, MobileNetV2 proposed by Sandlers *et. al.* [11] (2019). This scheme identifies these diseases with almost 90% accuracy. A common type of kidney cancer is renal cell carcinoma (RCC) which results into kidney disfunction. A method for treatment of RCC proposed by Chenjie *et. al.* [12] (2019). This method used tree ensemble method for extraction and identification of contributing features. For result interpretation they used shapely additive explanations (SHAP) XAI method to plot a decision curve analysis for clinical evaluation.

A UNet-PWP with gradient-weighted class activation mapping (GCAM)-Attention Fusion model for segmentation of kidney tumor proposed by Kiran Rao *et. al.* [13] (2023). The proposed architecture used UNet-PWP for precise kidney tumor segmentation. Segmentation results interpreted by attention and Grad-CAM XAI methods. Another scheme used lightweight convolutional neural network (CNN) for identification of kidney ailments (cyst, tumor, stone) and normal kidney image proposed by Bhandari *et. al.* [14] (2023). Results of proposed method explained using SHAP and LIME methods. Random forest (RF) classifier identified images in normal and abnormal (CKD) class with 98% accuracy. A CNN based method for identification of Normal kidney image and different kidney ailments (cyst, tumor stone) proposed by Ayub *et. al.* [15] (2024).

The proposed method augmented with synthetic minority over-sampling technique (SMOTE) for balanced dataset and Grad-CAM for result interpretability. Autosomal dominant polycystic kidney disease (ADPKD) is a subset disease of CKD, leads to end-stage renal disease (ESRD). For determination of ADPKD progression, total kidney volume (TKV) estimated from kidney image. Estimation of TKV is a time-consuming process and requires expertise, to produce result. To understand the inflation of kidney and track the progression of CKD, Dwiyanti *et. al.* [16] (2024) proposed a XGBoost model with the Synthetic SMOTE scheme. For classification of CKD, support vector machine (SVM), decision tree (DT), random forest (RF), k-nearest neighbors (KNN), XG-Boost, and naïve Bayes (NB) classifiers trained and tested on 340 US images. Kidney cyst, kidney stone and kidney tumor are considered as a precursor to CKD [17]. These ailments lead to kidney disfunction and results into any CKD stage [17]. Identification of Kidney cyst, stone and tumor from CT images using ViT with XAI ensemble modeling proposed by Arifuzzaman *et. al.* [17] (2024). This scheme identifies these precursors of CKD with 96% accuracy. All the results integrated through an ensemble technique which provide 96% detection accuracy. Another scheme using CNN with Grad-CAM based XAI model proposed by Tejaswini *et. al.* [18] (2025). This scheme identifies these precursors of CKD with 98% accuracy. A HybridResNetEffNet model which is mix of ResNet-50 and EfficientNet-B0 models proposed by Magna Benita *et. al.* [19] (2025). In this work they considered CKD stage-0, stage-1, stage-2, stage-3, stage-4. To deal with class imbalance and boost generalizability, proposed model incorporated stratified five-fold cross-validation, weighted loss functions, and ensemble voting. Their proposed model identifies individual CKD stages with 95% accuracy. XAI method, SHAP used to improve model transparency by Fizhan *et. al.* [20] (2025). For identification of patients belong to CKD Stage-3, Stage-4 and Stage-5 and other than these CKD stages, a scheme proposed by Nermeen *et. al.* [21] (2025).

This method used a generative adversarial network (GAN) to handle missing images in CKD datasets, few-shot learning, clubbed with explainable machine learning (XML) for CKD prediction. ML models such as SVM, logistic regression (LR), DT, RF, and voting ensemble learning (VEL), used for result comparison. For result interpretation techniques of XAI, SHAP and LIME, used. Another scheme mix of Catboost and LightGBM models for identification of CKD in CKD and Non-CKD proposed by Safiya Anjum *et. al.* [22] (2025). For explainability in ultrasound image interpretation, Grad-CAM is combined with SHAP and counterfactual analysis. This method identifies images in CKD and Non-CKD state with 99% accuracy.

From above literature review, it is observed that, [23] used indirect method for identification of amount of renal impairment, [13] used a method for segmentation and XAI for result interpretation, [14], [24], and [15] used different algorithms to find the the pre-conditions which leads to CKD. Whereas [16] used kidney volume for CKD stage progression, which require huge volume of data for each CKD stage, [25] classified CKD images into CKD stage-0, stage-1, stage-2, stage-3, stage-4, but the classification accuracy is less, [20] used algorithms with XAI for CKD classification and [21] used more than two algorithms for identification of CKD Stage-3,4,5.

A method is required which can identify individual CKD stage from the kidney images. Classify stage wise CKD affected images using one algorithm with less computational resources. Stage wise result interpretation using XAI algorithm for each CKD stage.

To deal with all above stated research gaps, we have devised a scheme which can identify the images of CKD stage along with normal kidney images with minimal computational resources and for interpretation of results used XAI method. The contributions in this work are meticulously curated CKD stage wise and normal kidney image datasets. Ultrasound (US) kidney images sorted, labelled and segmented for precise kidney area, under the supervision of experienced radiologist. An ensemble Vision Transformer framework combined with explainable AI techniques (LIME, LRP, and attention-based explanations) developed to achieve robust CKD stage classification while enhancing model transparency, interpretability, and clinical trust. Classification and interpretation of Normal kidney and five CKD stages.

The remaining part of the paper is organized as follow: Section 2 presents the material and method for CKD stage identification and XAI implementation. Section 3 describes the experimental setup. In this section, details of dataset, tuning details of selected models and other details about XAI are discussed. Section 4 explore the results obtained and discussion on performance of selected model and XAI method for interpretability. In this section also, the obtained results compared with similar existing methods. Finally, the paper ends with few concluding remarks on the performance of best model in Section 5.

## 2 Materials and Method

This section discusses about the workflow diagram for the proposed work and the dataset used.

Figure 1 shows a workflow diagram for CKD stage classification and XAI implemented in this work.

The collection of US kidney image data set is a crucial task. US kidney images collected, labelled and segregated into normal and five CKD classes with the help of team of experienced radiologist. The labelled dataset is imbalanced in nature. To avoid any biasing in the classification results and subsequent interpretation of results synthetic minority over-sampling technique (SMOTE) with fine-grained data augmentation used to generate synthetic samples with enhance intra-class variability and robustness. The patchifying block transform the input into tokens. These tokens further processed by Vision Transformer (ViT). ViT cannot process directly the full image like CNN's, instead, they process image patches as sequential tokens. The patch embedding layer perform the vector embedding on transformed input. The input tensor undergoes transformations in each layer of ViT, affecting both its values and shape.

When the tensor reaches the embedding layer, its shape becomes [1,4,768] × [1,4,768], where 768 represents the embedding dimension.

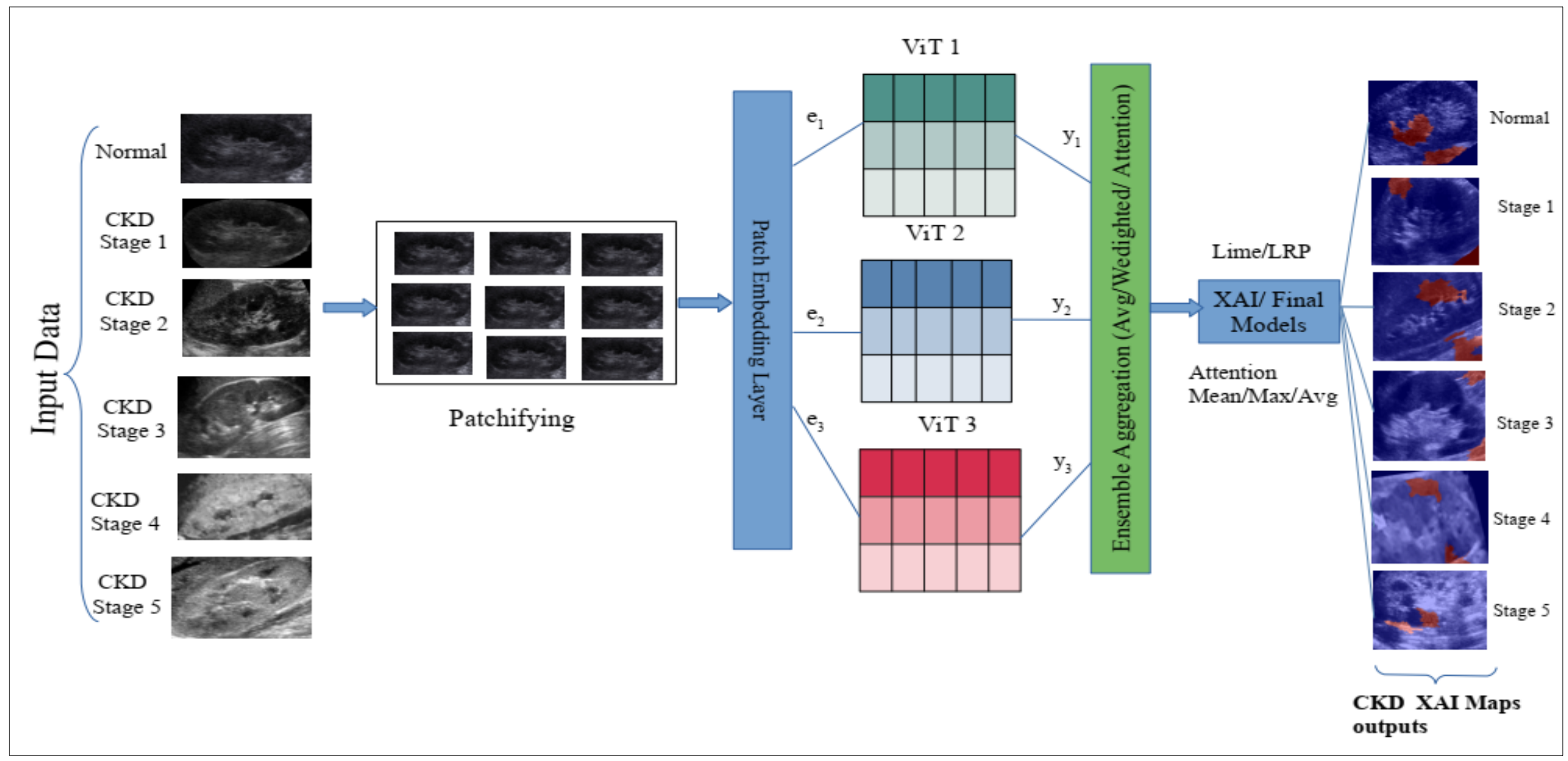

Figure-1: CKD Stage classification and XAI workflow diagram.

The embedding layer converts each image patch into a dense numerical feature vector that can be processed by the ViT. US CKD images divided into small patches, the embedding layer transforms these patches into meaningful representations by capturing visual information such as texture, intensity, and structural patterns. These embedded vectors act as input tokens for the transformer, enabling it to learn relationships between different regions of the kidney image for accurate CKD stage classification. The input tensor undergoes transformations in each layer, affecting both its values and, in some cases, its shape.

The ensemble aggregation layer combines outputs from multiple ViT's to generate accurate and reliable CKD prediction. It uses methods like average, weighted, or attention-based aggregation to merge the learned features from different models. The ensemble aggregation layer improves the accuracy, robustness, and reliability of CKD stage classification by combining the strengths of multiple ViT models and reducing prediction errors.

**3 Experimental Setup**

To validate the performance of the proposed XEns-CKD framework, a series of experiments were conducted under a standardized experimental setting. The details of the dataset, preprocessing procedures, model configuration, and evaluation criteria are described in the following subsections.

### 3.1 Dataset

Open access US CKD image dataset is not available. Curation of US CKD image dataset is an important and challenging task of proposed method. With approval from competent authorities and patient's consent in writing, following ethical and professional norms, US CKD stagewise images and normal kidney images collected at radiology center Shivam diagnostic, Jalgaon, Maharashtra, India. Under the supervision of two radiologist with 15 years of experience, this dataset collected, sorted, labelled for CKD stages and normal kidney images.

For sorting, identification and classification of images belongs to CKD stages and Normal state of kidney, of Dr. Aleem Ansari (Medical Officer at District Civil Hospital, Dhule, Maharashtra, India) guided and helped. Figure-2 shows the US normal and CKD images belonging to different stages. It is seen from images that, progression of disease from normal to stage-1 to stage-5 results into shape deformation and increment in gray values of the pixels. Features representing change in pixel gray values and shape deformation if captured and processed can be helpful in CKD classification.

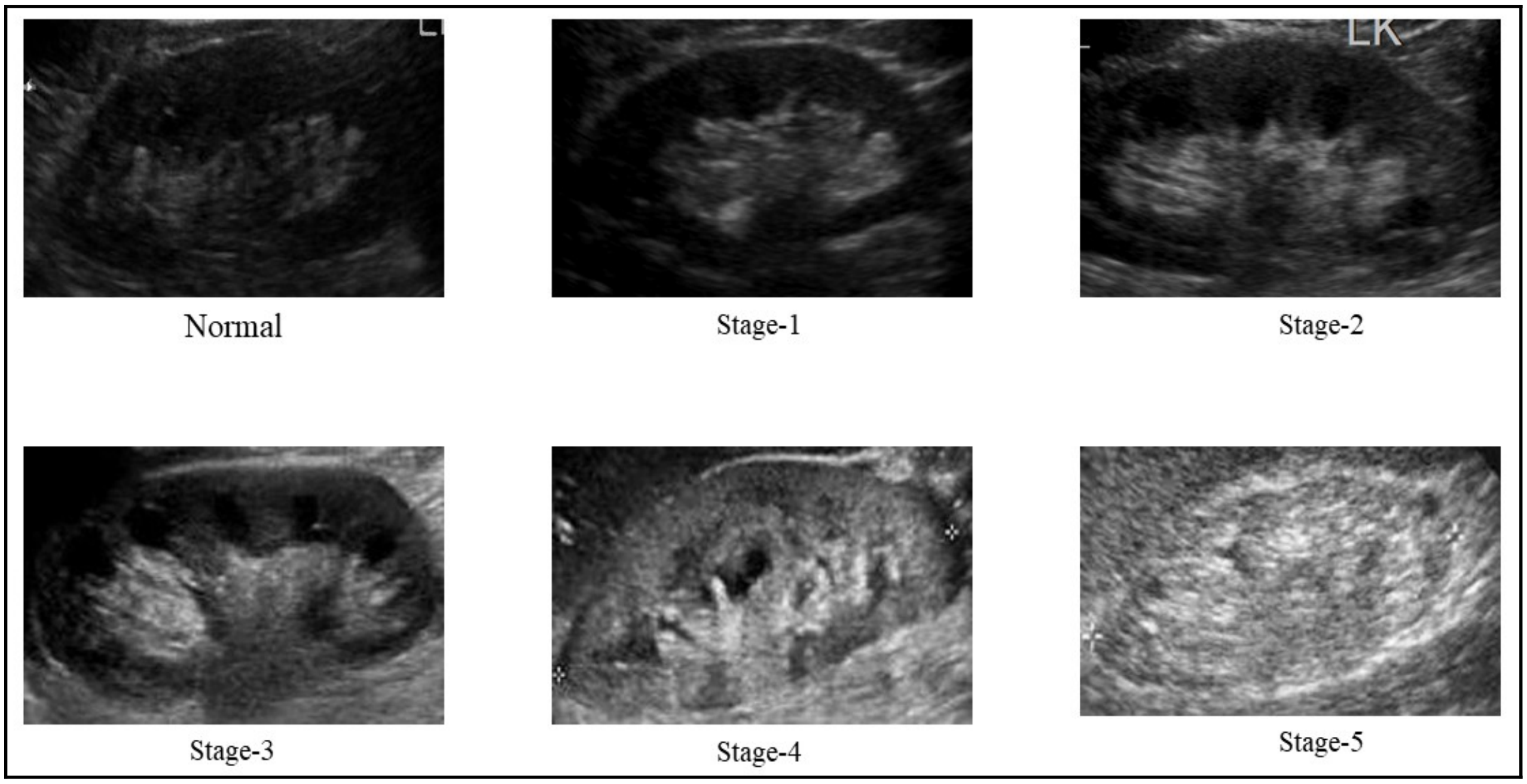

Figure-2: Images of Normal and different CKD Stages.

The collected dataset has imbalance for images belong to CKD stage-4 and CKD stage-5. This is because of the critical stage of kidney. Once the kidney reaches this stage, usually clinicians suggest for GFR test. Therefore, images received from radiology center are less for CKD stage-4 and CKD stage-5. To avoid biasing in the classification result synthetic minority over-sampling technique (SMOTE) [26], with fine-grained data augmentation used to generate balanced synthetic samples with enhanced intra-class variability and robustness. Also, SMOTE has been tuned to perform under sampling for class normal. Table 1 shows the number of CKD stagewise images before and after the SMOTE operation. The balanced US kidney image dataset presented in Table 1 is used for training, testing and XAI process.

Table 1: Comparison before and after SMOTE.

| **CKD Class** | **Number of images** | |
|---|---|---|
| | **Before** | **After** |
| Normal | 107 | 100 |
| Stage 1 | 84 | 100 |
| Stage 2 | 93 | 100 |
| Stage 3 | 73 | 100 |
| Stage 4 | 29 | 100 |
| Stage 5 | 29 | 100 |

**3.3 Model Tuning:**

Vision Transformer (ViT) model known for adapting the transformer architecture from natural language processing (NLP) to computer vision by processing image patches via self-attention [27]. Figure 3 illustrates the overall workflow of the proposed CKD stage classification framework. First, the input US kidney images are processed independently by three Vision Transformer (ViT) models with different configurations to learn diverse feature representations. The predictions generated by these models are then combined using an ensemble aggregation strategy (e.g., SoftMax probability averaging, weighted averaging, or voting) to produce a more robust and accurate final CKD stage classification result.

Finally, the ensemble prediction interpreted using Explainable Artificial Intelligence (XAI) techniques, including LIME, Layer-wise Relevance Propagation (LRP), and Attention-based visualization methods, which highlight the image regions contributing to the model's decision, thereby improving transparency, interpretability, and clinical trust in the classification results.

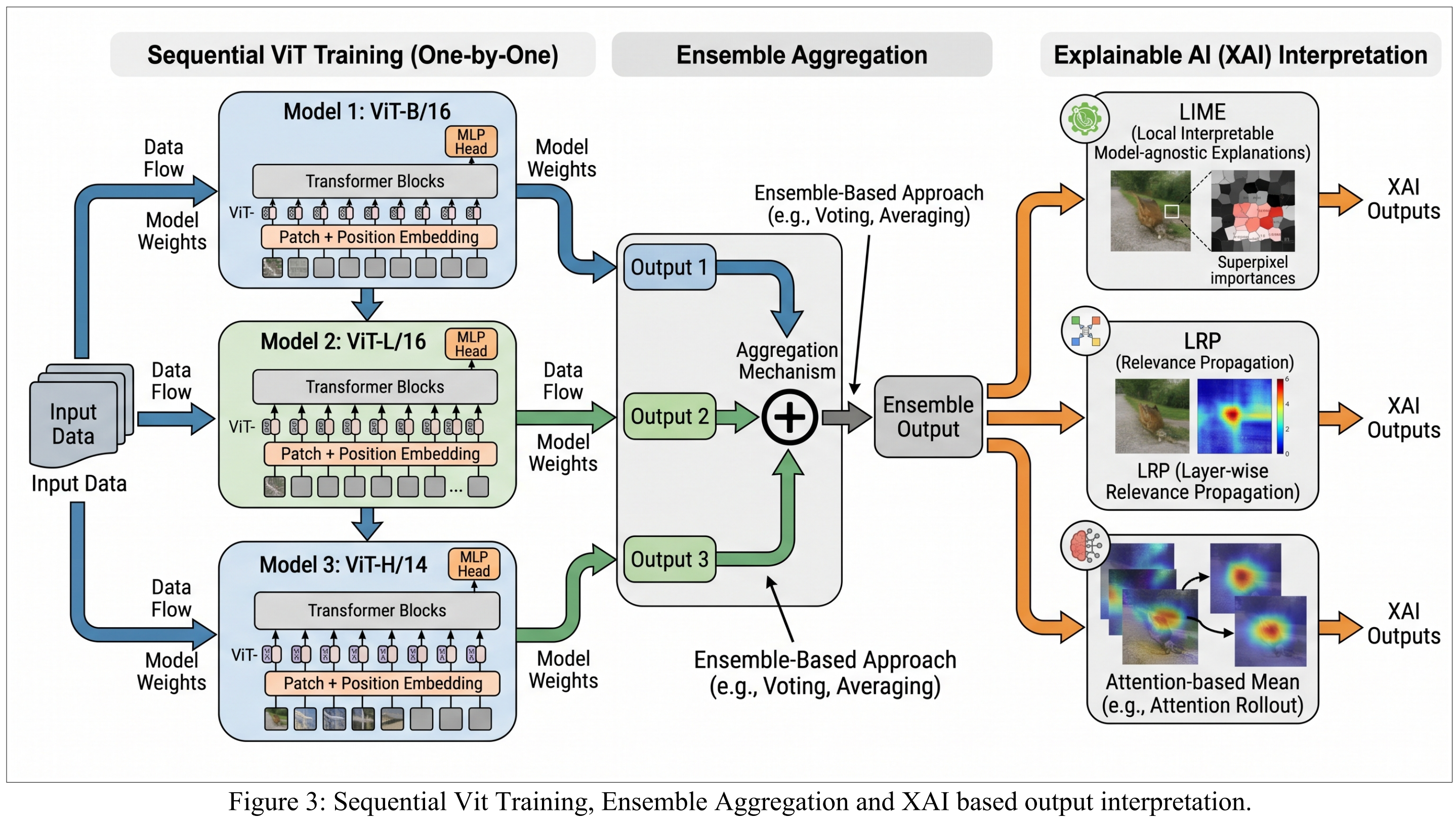

Figure 3: Sequential Vit Training, Ensemble Aggregation and XAI based output interpretation.

### 3.4 Implementation Details and Mathematical Formulation

All three ViT models were trained independently on an NVIDIA A100 GPU using identical hyperparameter settings to ensure a fair comparison. The complete training configuration is provided in Table 2. Let I denote an input kidney ultrasound image. Each image is divided into N non-overlapping patches. The patches are transformed into embedded tokens and combined with positional encodings before being processed by the ViT : $z_i = E(P_i) + E_{pos}(i), i = 1, \cdots, N$ where E(Pi) represents the patch embedding operation and Epos(i) denotes positional encoding.

The three constituent ViT models independently extract discriminative feature representations from the kidney ultrasound images. For each model, the output logits are converted into a class-probability vector using the SoftMax function: $p_k = SoftMax(f_k(I)), k = 1, 2, 3,$ where fk(I) represents the output of the kth Vision Transformer model and pk denotes the corresponding class probability vector.

To improve robustness and reduce model-specific bias, the outputs of all three Vision Transformers are combined using SoftMax probability averaging. $p_{ens} = \frac{1}{3}\sum_{k=1}^{3} p_k$ , where where pens denotes the ensemble class-probability vector. The final CKD stage prediction is obtained by selecting the class with the highest ensemble probability. $\hat{c} = \arg\max_{c \in C} p_{ens}(c)$ where c represents one of the six classes including Normal kidney and CKD Stages 1–5.

To enhance model transparency and facilitate clinical interpretation, three Explainable Artificial Intelligence techniques LIME, Layer-wise Relevance Propagation (LRP), and attention-based visualization are applied to interpret the ensemble prediction. In general, the explanation map generated by an XAI method can be expressed as: $M_j = Phi_j(I, \hat{c}, F_{ens})$, where $M_j$ denotes the explanation map generated by the $j_{th}$ XAI method, $\Phi_j$ represents the corresponding explanation function, and Fens is the ensemble model.

For LRP, the prediction score is propagated backward through the network while approximately preserving the total relevance: $\sum_i R_i \approx f_{ens,\hat{c}}(I)$, where where $R_i$ represents the relevance assigned to the $i_{th}$ input feature and fens,c^(I) denotes the ensemble score for the predicted class. For attention-based interpretation, attention maps obtained from the L transformer layers are aggregated to identify the image regions that contribute most strongly to the prediction: $A_{agg} = \frac{1}{L} \sum_{l=1}^{L} A^{(l)}$, where where A(l) represents the attention map obtained from the lth transformer layer and Aagg is the aggregated attention map.

The resulting LIME, LRP, and attention-based maps provide complementary visual explanations of the image regions influencing CKD-stage classification. These explanations enhance the transparency and interpretability of the proposed ensemble ViT framework and support its potential application in clinical decision-support systems.

Table 2: Details of hyper parameters used for ViT's tuning.

| Model | Learning Rate | No. of Freeze layer | Schedulers | weight decay |
|---|---|---|---|---|
| **ViT1** | 0.0003 | 6 | Cosine | 0.01 |
| **ViT2** | 0.0002 | 8 | Plateau | 0.01 |
| **ViT3** | 0.0005 | 4 | Cosine | 0.05 |

### 3.5 Evaluation Metrics

The classification performance of each model was evaluated using accuracy, sensitivity, specificity, precision, F1-score, Matthews correlation coefficient (MCC), Youden's index, and the area under the receiver operating characteristic curve (AUC). Accuracy represents the proportion of correctly classified samples and was calculated as Accuracy = (TP + TN)/(TP + TN + FP + FN). Sensitivity, also known as recall or the true-positive rate, represents the proportion of actual positive samples correctly identified and was calculated as Sensitivity = TP/(TP + FN). Specificity, or the true-negative rate, represents the proportion of actual negative samples correctly identified and was calculated as Specificity = TN/(TN + FP). Precision represents the proportion of predicted positive samples correctly classified and was calculated as Precision = TP/(TP + FP).

The F1-score is the harmonic mean of precision and sensitivity and was calculated as F1-score = (2 × Precision × Sensitivity)/(Precision + Sensitivity). MCC was calculated as MCC = (TP × TN − FP × FN)/√[(TP + FP)(TP + FN)(TN + FP)(TN + FN)]. Youden's index was calculated as J = Sensitivity + Specificity − 1. Finally, AUC was calculated from the ROC curve using predicted class probabilities to assess the ability of each model to discriminate among classes across different decision thresholds. For the six-class classification problem, class-specific metrics were calculated using a one-versus-rest approach and summarized using macro-averaged and weighted-averaged values.

## 4 Results and Discussions

This section presents a comprehensive evaluation of the proposed **XEns-CKD** framework through both quantitative and qualitative analyses. The performance of the proposed method is compared with existing state-of-the-art approaches using standard evaluation metrics to demonstrate its effectiveness and reliability.

### 4.1 Qualitative Results

A single Vision Transformer (ViT) model can provide accurate and reliable classification results, but its performance may be influenced by the model-specific biases and variations. So, to reduce dependence on a single model and improve robustness in predictions, we used an ensemble framework which used three independent ViT models. Rather than relying on the prediction of a single network, the outputs of all three models are integrated through an ensemble strategy to obtain the final classification result.

This approach minimizes the potential bias in model selection, which enhances the generalization capability, and improves classification stability across different CKD stages.

The ensemble method combines all the three results using a softmax probability averaging, which is efficient and different from accuracy averaging.

For quantifiable measure of classifier performance, parameters TP, TN, FP and FN estimated from confusion matrices. The ensemble results of ViT models are shown in Table-3. The other performance indicators such as recall, F1-score, and AUC score, are impressive for our ensemble model than baseline models, shown in Table-3.

Table-3: Parametric Evaluation of Individual Model.

| Model | Accuracy | MCC | Macro Sensitivit y | Macro Specificity | Macro Precision | Macro F1-score | Macro Youden Index | Macro Balanced Accuracy |
|---|---|---|---|---|---|---|---|---|
| **ViT1** | 84.848 | 0.8186 | 0.8485 | 0.9697 | 0.848 | 0.847 | 0.8182 | 0.9091 |
| **ViT2** | 83.33 | 0.8051 | 0.8333 | 0.9667 | 0.8612 | 0.8308 | 0.8 | 0.90 |
| **ViT3** | 72.72 | 0.6765 | 0.7273 | 0.9455 | 0.7291 | 0.7177 | 0.6727 | 0.8364 |

**Legends: Sens.:** Sensitivity**, Sp.:** Specificity**, Pr.:** Precision**, MCC:** Mathews Correlation Coefficient**, YI:** Youden Index**.**

Figure 4(a) to Figure 4(h) presents a comparative parametric evaluation of the proposed Ensemble ViT model against the individual Vision Transformer models (ViT1, ViT2, and ViT3) across multiple performance metrics, including Accuracy, MCC, Macro Sensitivity, Macro Specificity, Macro Precision, Macro F1-score, Macro Youden Index, and Macro Balanced Accuracy.

Figure 4a show the classification accuracy achieved by ViT1, ViT2, ViT3, and the proposed Ensemble model. The Ensemble model achieved the highest accuracy of 86.36%, which is outperforming ViT1 (84.85%), ViT2 (83.33%), and ViT3 (72.72%). This shows that combination of predictions from multiple ViT's can improves the overall classification performance by reducing the individual model errors and increasing the prediction reliability of individual models.

Figure 4b show the Matthews Correlation Coefficient (MCC), which is able to helps to evaluates the classification quality of model by considering true positives, true negatives, false positives, and false negatives. The proposed Ensemble model achieved the highest MCC of 0.8375, which is higher than other by ViT1 (0.8186), ViT2 (0.8051), and ViT3 (0.6765). The higher MCC points that the Ensemble model provides more balanced and reliable predictions across all classes among all models.

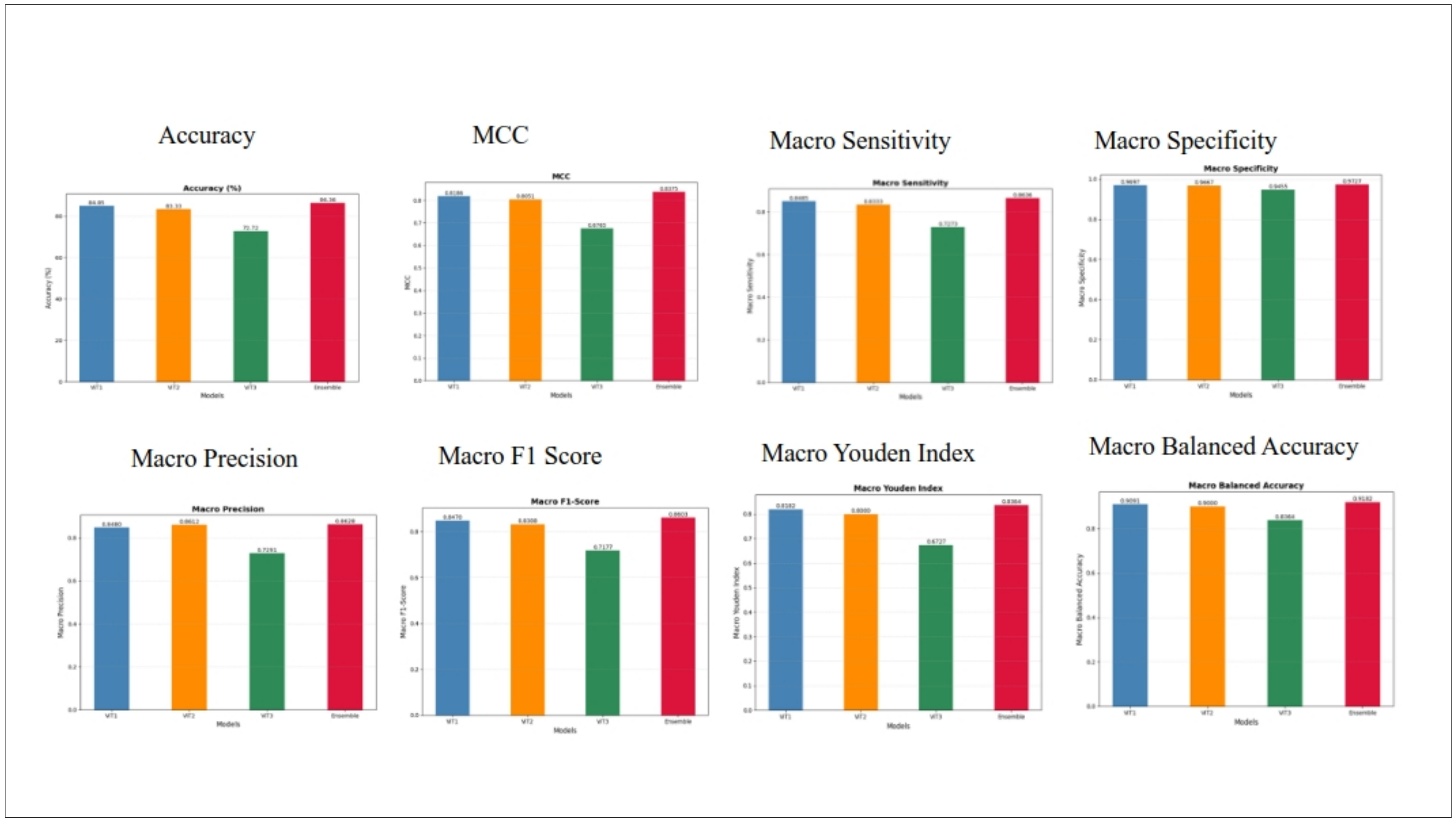

Figure 4: Performance of ViT1, ViT2, ViT3, and the proposed Ensemble ViT model across multiple classification metrics for CKD stage prediction.

Figure 4c shows the Macro Sensitivity also knows as Recall for all models. The Ensemble model get the highest macro sensitivity which is 0.8636, indicating towards the superior capability of model in the correctively identifying samples from every class. ViT1 and ViT2 achieved 0.8485 and 0.8333, respectively, whereas ViT3 recorded the lowest value (0.7272). The improved recall shows the effectiveness of ensemble learning in the minimizing false negatives.

Figure 4d demonstrate the results of Macro Specificity metric. In this, Ensemble model achieved the highest specificity equals to 0.9727, which indicates towards the excellent capability of it in correctively rejecting negative samples. ViT1 (0.9697) and ViT2 (0.9667) also performed very well, while ViT3 achieved 0.9455. This high specificity confirms that the proposed model able to maintains a very low false-positive rate.

Figure 4e shows Macro Precision metric values. In this, Ensemble model achieved the highest precision of 0.8628, which is close to ViT2 (0.8612) and ViT1 (0.8480), whereas ViT3 obtained the lowest in all, which equals to 0.7291. This higher precision helps to determine that the Ensemble model generates fewer false-positive predictions while maintaining the strong classification performance.

Figure 4f shows t Macro F1-Score metric scores. In this, Ensemble model achieved the highest F1-score of 0.8603, so it performing better than ViT1 (0.8470), ViT2 (0.8308), and ViT3 (0.7177). This shows that the Ensemble model can provides the best trade-off between correctly identifying positive samples and minimizing false predictions.

Figure 4g shows the Macro Youden Index scores of models, which helps to measures the diagnostic effectiveness of the classifiers by combining sensitivity and specificity. In which, the Ensemble model achieved the highest Youden Index of 0.8364, for ViT1, ViT2, and ViT3 scores are 0.8182, 0.8000 and 0.6727 respectively. The higher index shows that the proposed Ensemble model offers the best discriminative capability among all evaluated methods.

Figure 4h shows the Macro Balanced Accuracy scores of all models. In this metric, the Ensemble model achieved the highest balanced accuracy of 0.9182, which is higher than the other such as ViT1 (0.9091), ViT2 (0.9000), and ViT3 (0.8364). This result shows that the Ensemble model maintains consistent performance across all classes, which is making it particularly suitable model for the multiclass medical image classification tasks where class imbalance may exist.

From Figure-4 it is clear that after ensemble of all three vits using SoftMax, we have ensemble model accuracy as 86.36%. This shows the diversity among all the models which are contributing in the performance improving. The confusion matrices of the individual models (ViT1, ViT2, ViT3) and the proposed ensemble model are shown in Figure 5a to Figure 5d. These confusion matrices provide a detailed class-wise evaluation of the classification performance across the six categories (Normal and CKD Stages 1–5).

For VIT1 (Accuracy: 84.85%), the confusion matrix shows the strong diagonal dominance, which indicating generally reliable classification. However, minor misclassification is observed between the Normal and Stage 1 (1 instance), which suggesting the overlap in early-stage features. More importantly, Stage 2 exhibits confusion with Stage 3 (3 instances), which reflecting difficulty in distinguishing between the intermediate CKD stages where structural differences are subtle. A small degree of misclassification is also observed from Stage 3 to Stage 4, which is also indicating transitional ambiguity between adjacent disease stages.

For VIT2 (Accuracy: 83.33%), the misclassification pattern becomes more pronounced compared to previous one. The confusion between Normal and Stage 1 increases (2 instances), indicating the reduced sensitivity to early-stage distinctions between classes. Additionally, Stage 2 shows the significant confusion in the Stage 3 (5 misclassified vs 6 correctly classified), which is highlighting the instability in mid-stage prediction of the model. While later stages such as Stage 4 and Stage 5 are classified with the relatively high accuracy, but the inconsistency in earlier and intermediate stages reduces overall reliability of the model.

For VIT3 (Accuracy: 72.72%), the confusion matrix shows the large amount of degradation in the performance. Notably, Stage 1 is misclassified as Normal (3 instances), which indicating the poor discrimination quality of early disease onset. Furthermore, Stage 3 exhibits widespread confusion with the Stage 2 and Stage 4, and Stage 4 is misclassified as Stage 5 (4 instances). These results suggest that VIT3 fails to learn the stable and discriminative representations, particularly for transitional and clinically critical stages.

### 4.2 Ensemble Model Analysis

The ensemble model which has accuracy score of 86.36% shows an improvement in the classification behaviour of the model. The strong diagonal dominance across all classes indicates the improved true positive rates. Near-perfect classification for Stage 1, Stage 4, and Stage 5, with the negligible or no misclassification makes it more reliable compared to other. Reduced values of confusion matrices between Stage 2 and Stage 3, which were the most challenging classes for the detection by individual models. Overall, the reduction in the off-diagonal errors, indicating towards the improved class separability.

To further analyze all model behaviour, explainability methods were applied on the three individual Vision Transformer models (ViT 1, ViT 2, and ViT 3). Figures 6(a) to 6(c) present the XAI results (LIME, LRP, and Attention) across all CKD stages for each model. From Figures 6(a) to 6(c) of all three models it is observed that: LIME was mainly designed for interpretation of local area [29]. LIME produces sparse and localized highlights but lacks consistency across samples and stages. This is mainly because of obtained results of CKD and Normal kidney images. The texture variation in normal to CKD stage-1 to stage-2 are minute. As the number of healthy nephrons at these stages are more. At Stage-3 and particularly at Stage-4 and Stage-5, the healthy nephrons decline is drastic.

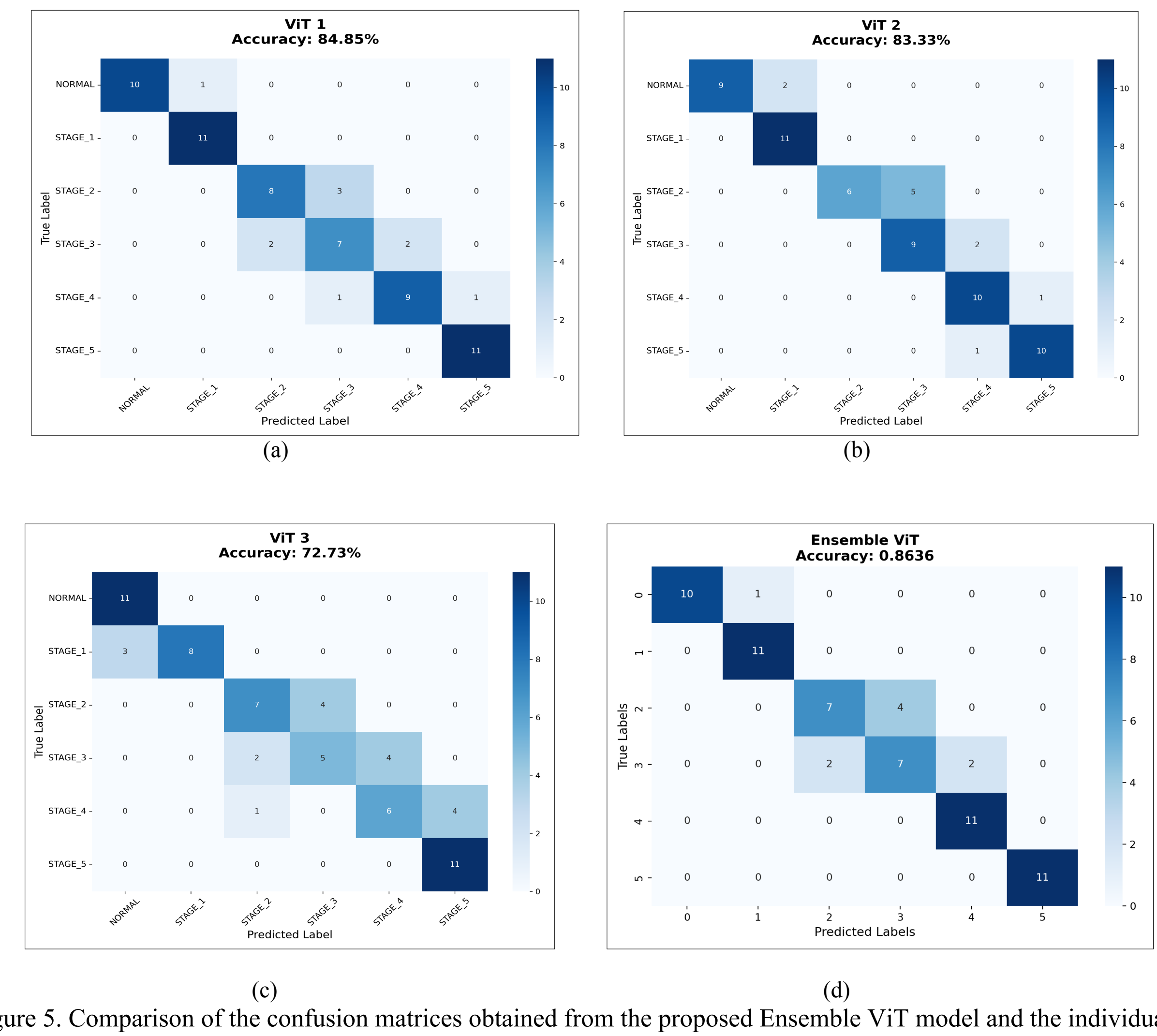

Figure 5. Comparison of the confusion matrices obtained from the proposed Ensemble ViT model and the individual Vision Transformer models (ViT1, ViT2, and ViT3) for CKD stage classification.

Due to which LIME unable to produce a consistency in result for these stages. The texture variations in CKD Stage-3 to Stage-5 are visible. Therefore, LIME able to identify the local areas contributing in progression of CKD.

The objective of LRP is to establish a relevance between neurons of output layer and input layer [30]. In this CKD stage classification task, the number of neurons involved in classification task vary for Normal, and CKD stage-1 to Stage-5. Because of the variations in number of nephrons, texture and grey values of pixels, even though LRP captures broader regions; however, the explanations are often noisy and less focused as expected.

Attention maps used for interpretation pinpoint the areas contributing in CKD progression.

Particularly shown in Fig (6.a) and Fig (6.b), Attention map highlights the area contributing in decision to identify the reason for CKD progression. Majority of nephrons found in Medulla (the central part) and Cortex (the outer part) regions of human kidney. As highlighted in these two Figures, most of the nephrons died due to CKD progression. Attention maps are comparatively smoother than others, but remain diffuse and less sharply localized, especially in early stages.
All these interpretations align with clinical observations and verified by the Dr. Ansari. Across disease progression, all models show difficulty in early stages (Normal, Stage 1 and Stage 2), where subtle structural differences limit clear feature attribution. In contrast, later stages (Stage 4 and stage 5) show more concentrated and meaningful highlighted regions.

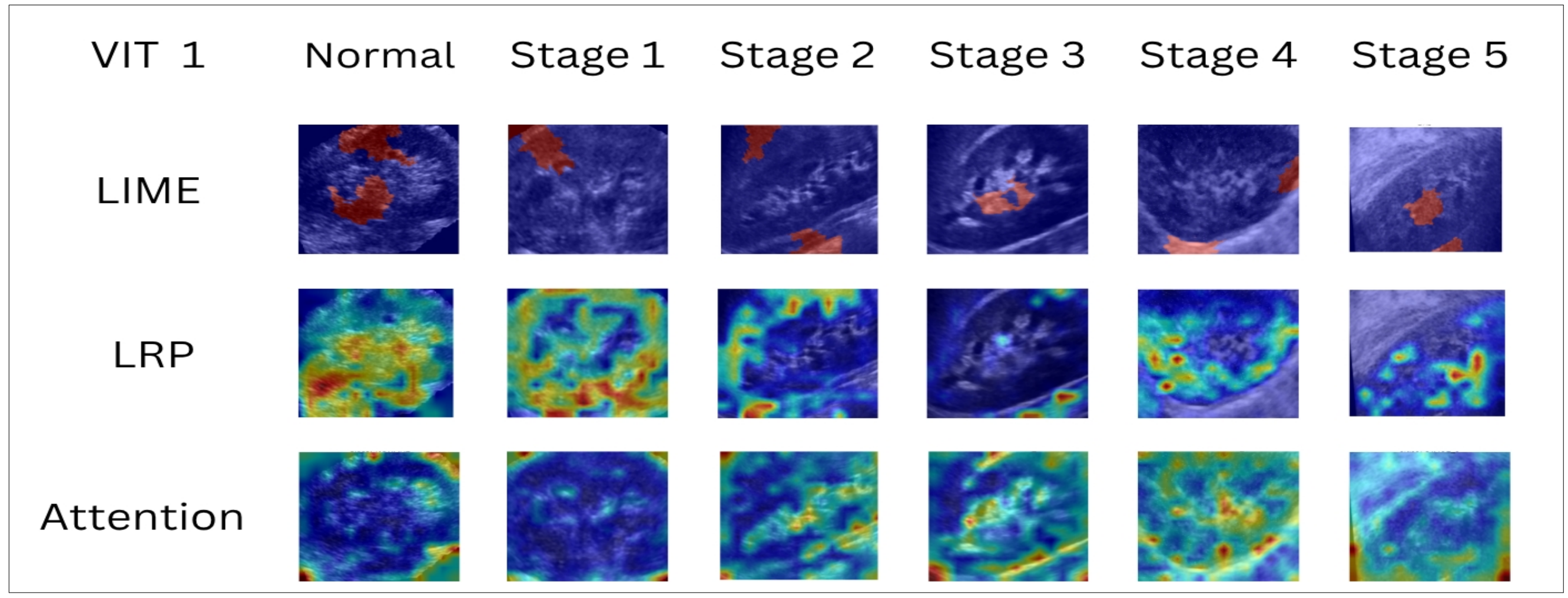

(6.a) ViT 1 XAI results

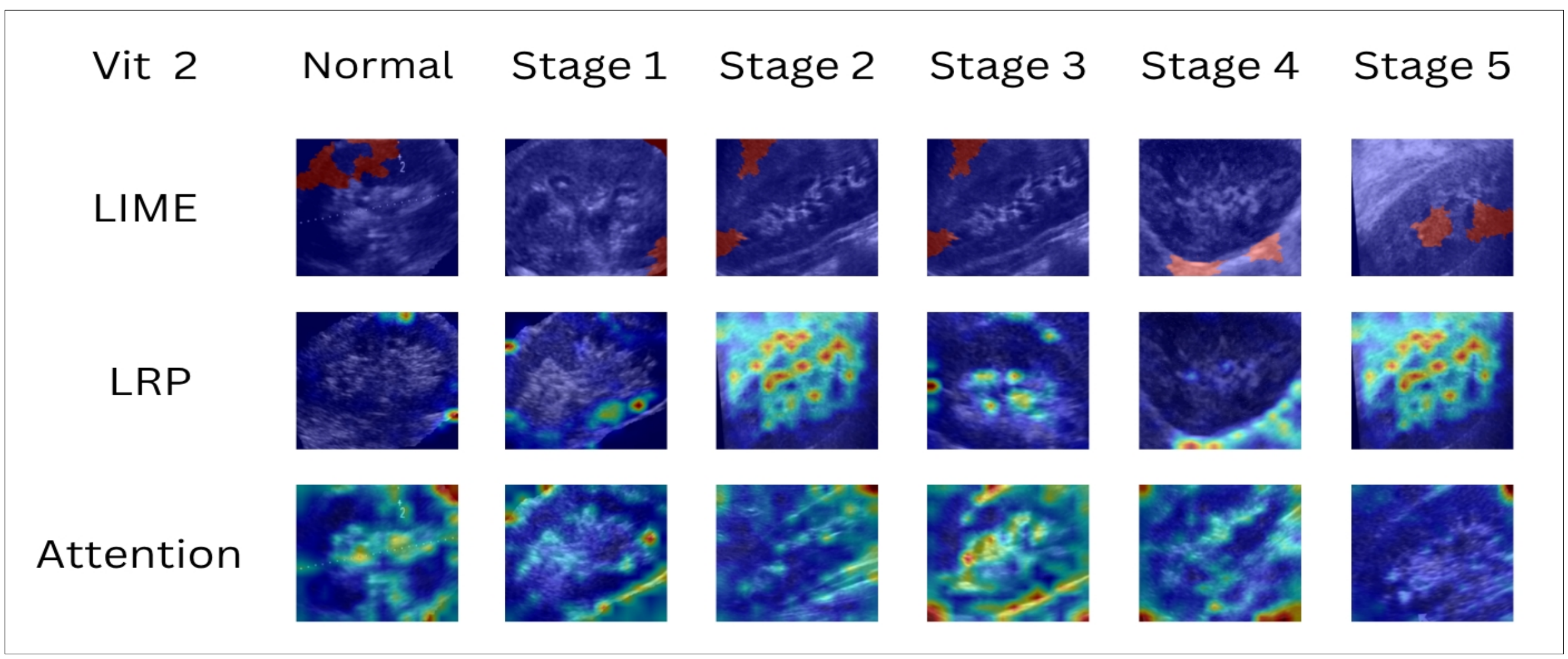

(6.b) ViT 2 XAI results

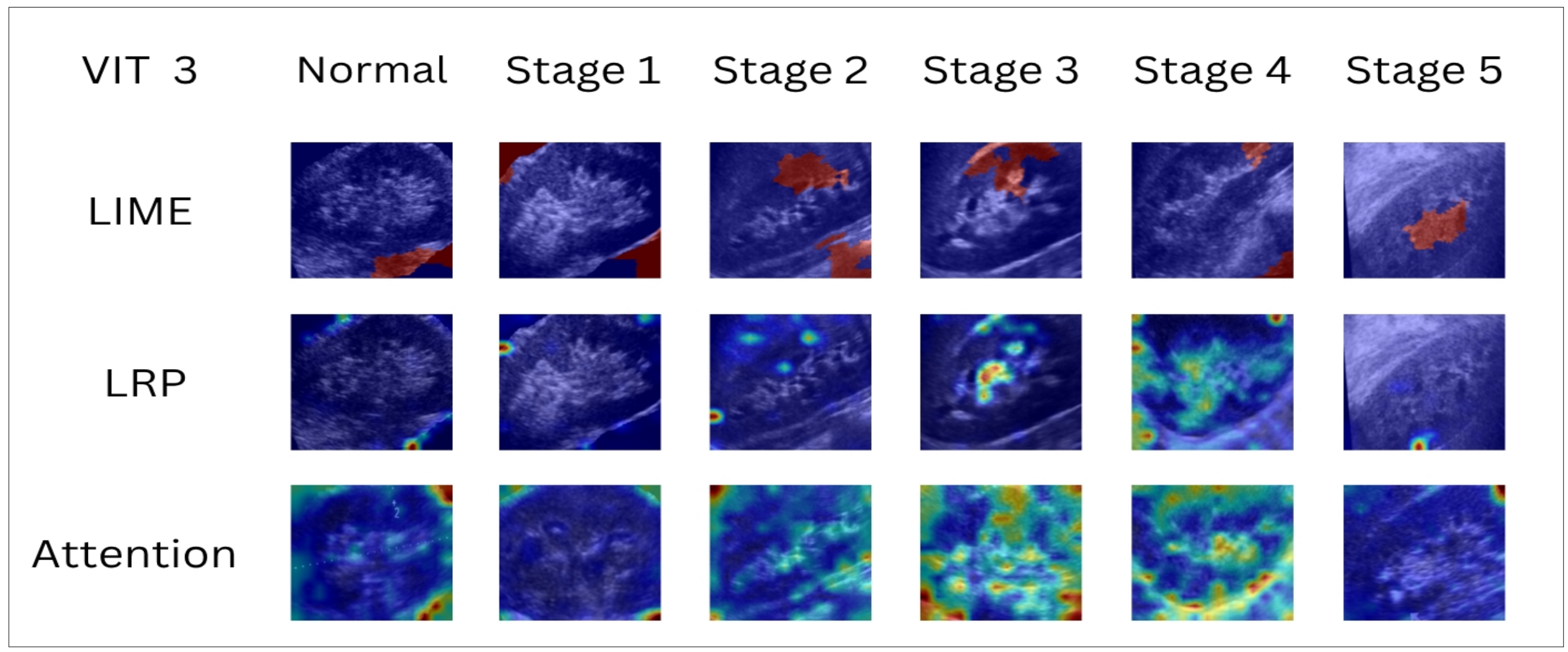

(6.c) ViT 3 XAI results

Figure 6: XAI visualizations produced by ViT-1, ViT-2, ViT-3 for CKD classification. Rows represent CKD stages (Normal, Stage 1–Stage 5), while columns correspond to LIME, LRP, and Attention Map explanations.

While the overall behaviour of all models is consistent but slight variations exist between models (Figures 6(a) to figure 6(c)), with differences in the location and spread of highlighted regions. This indicates that individual models learn slightly different feature representations.

When compared to the ensemble model (Figure 7), all three individual ViT models shows: Less stable attention patterns. Greater variability in highlighted regions. Lower consistency across samples. Overall, although individual models can ably identify relevant features but their explanations are less reliable and consistent compared than the results produced by the ensemble model.

The explainability analysis indicates that the explanation quality may vary depending on the complexity of class features. In medical imaging tasks such as CKD staging: Early stages (Stage 1 and 2) often have very small differences which makes explanation generation more challenging. Advanced stages (Stage 4 and 5) show clearer structural change which helps to enable more reliable and interpretable attributions, shown in Figure 7.

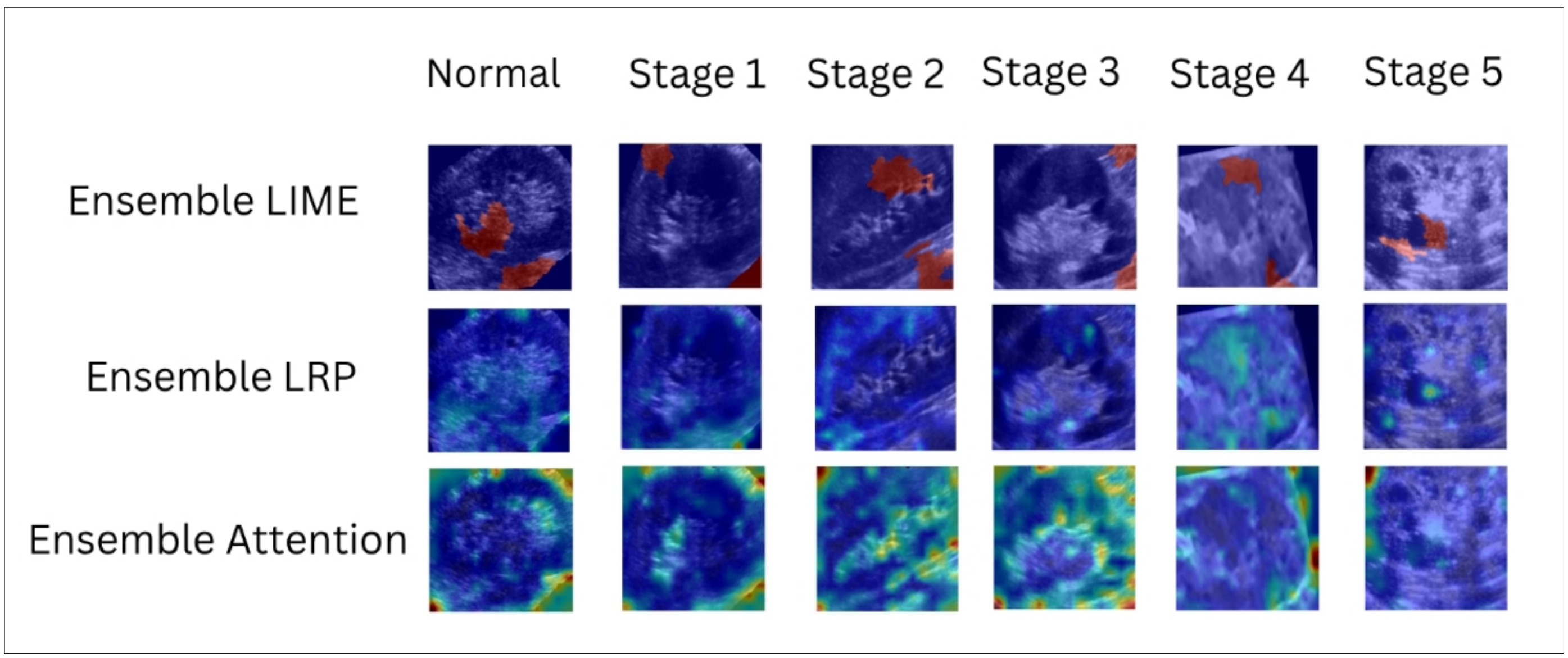

Figure 7: Ensemble ViT XAI results

**4.3 Quantitative Results**

Table-4 presents a comparative evaluation of different explainability methods used in this study across three key dimensions which are faithfulness, sensitivity, and complexity. Among the all evaluated approaches on models, Attention (Minimum) achieves the highest faithfulness score which is 0.0294, which indicating towards a relatively stronger alignment between highlighted regions and model predictions. LIME shows moderate faithfulness but perform best in maintaining the lowest complexity equals to 0.0897, which makes it more interpretable for human users.

In contrast, LRP exhibits slightly negative faithfulness values which is -0.0008, which suggest a weaker correlation with the prediction outcomes in this setup. However, its performance remains the competitive to others when considering sensitivity and robustness.

The sensitivity evaluates the stability of explanations under small changes in the input data. As shown in Table-4, Attention (Mean) produces the lowest sensitivity (0.041), which followed closely by Attention (Min) and Attention (Max). This indicates that attention-based methods generate more stable explanations when compared to the LRP and other methods. LRP records a comparatively higher sensitivity value which is 0.0754, which suggest that its explanations are more affected by the minor variations in input images. This may reduce its reliability in clinical scenarios because consistency is critical in medical scenarios.

Table-4: Ranking of XAI methods based on Faithfulness, Sensitivity, and Complexity metrics for CKD classification.

| Rank | Method | Faithfulness ↑ | Sensitivity ↓ | Complexity ↓ |
|---|---|---|---|---|
| **1** | Attention (Min) | 0.0294 | 0.0486 | 0.436 |
| **2** | LIME | 0.0048 | — | 0.0897 |
| **3** | LRP | -0.0008 | 0.0754 | 0.4751 |
| **4** | Attention (Mean) | -0.0012 | 0.041 | 0.9064 |
| **5** | Attention (Max) | -0.0084 | 0.0539 | 0.2015 |

An ensemble-based approach which consists of three Vision Transformer (ViT) models for the proposed framework used. The performance of the proposed ensemble model compared with the method of [18]. The experimental results show that the proposed approach achieves an improvement in classification accuracy for the CKD stage identification using ViT when compared with [31]. This enhancement can be attributed to the advantages of the proposed methodology, which includes the utilization of three ViT models and the aggregation of their predictions through an ensemble averaging strategy. Furthermore, unlike previous approaches, which has ovelookedd one or other stage of CKD or Normal Kidney state, proposed framework considers normal kidney images and all five stages of CKD, which enables comprehensive disease staging. The proposed model accurately able to classifies normal kidneys as well as CKD stages 1–5.

The ensemble framework achieved a weighted accuracy of 86.36%, shows its effectiveness and robustness in this work. In comparison with other baseline models.

Table-5 show the comparison of proposed work and existing works for CKD stage classification. Because of the very less work on CKD stage classification with XAI from US kidney images, works on same theme with different imaging modalities are also considered for comparison.

From Table-5, it is observed that, when the results of proposed work compared with [19] [20] and [31], it is observed that, the proposed method has considered Normal kidney stage and all Five stages of CKD. Even though he reported classification accuracy is low, but emphasize is on identification and interpretation of areas contributing in progression of CKD states. Our method also considered the US kidney images belong to Normal state. Identifying image belong to Normal state is also an important task, this will assist the clinician for better decision-making process.

Table-5: Evaluation metrics for our ensemble model and other state-of-the-art models.

| Imaging Modality | Classification in | Feature Extraction | Method | XAI with | Weighted Classification Accuracy (%) |
|---|---|---|---|---|---|
| **US** | CKD Stage-0, CKD Stage-1, CKD Stage-2, CKD Stage-3, CKD Stage-4, | ---- | Hybrid ResNetEffNet model [19] (2025) | GRAD-CAM | 95 |
| **US** | CKD, Non-CKD | GLCM,GLSZM, GLRLM,NGTDM, and GLDM | SVM [20] | SHAP | 94 |
| **US** | Precursors of CKD | ---- | ViT [31] | GRAD-CAM | 82 |
| **US** | Normal Kidney Image, CKD Stage-1, CKD Stage-2, CKD Stage-3, CKD Stage-4, CKD Stage-5. | ---- | Proposed (2026) | LIME, LRP, Attention Map with Ensemble Modeling | 86.36 |

When the performance of proposed model compared with [31] which deals with elements causing CKD, it is observed that, our method offers increment of 4% accuracy. This is mainly because of the factors such as use of real-world image dataset, tuning of ViT's and use of XAI techniques. Also in comparison with models used in [19] [31], our proposed model is free from limitations such as parameter redundancy, vanishing gradient problem, heavy memory constraints and hardware bottleneck issues. Another merit of our ViT ensemble method is, it can be integrated in mobile app with less resources.

## 5 Conclusion

CKD is a silent epidemic. Its progression not felt by the person. Classification of CKD can be done in five stages CKD Stage-1, Stage-2, Stage-3, Stage-4 and Stage-5. Where the Stage-5 is the last stage in which kidney fails to work. Early identification of CKD can save a person from financial and emotional burden. For CKD stage identification, we proposed a novel ensemble vision transformer (EViT) for classification of Normal and Five CKD stages of ultrasound images. Three ViT's are trained on private ultrasound CKD images dataset. Each ViT tuned on different training parameters. The performance of each ViT estimated using macro sensitivity, macro specificity, macro precision, macro F-1 score, macro Youden Index, MCC and macro balanced accuracy.

The ensemble models result demonstrate that overall classification accuracy is 86.36%. The obtained ensemble results interpreted using the explainable artificial intelligence (XAI) techniques layer-wise relevance propagation (LRP), Attention_Min, Attention_Max and LIME to enhance model transparency and clinical trust. The XAI method attention map identifies and interpret the CKD stage classification result with better confidence and trust. It correctly identifies the areas contributing in progression of CKD stages. When the results of proposed method compared with the existing schemes it is observed that, our method identifies CKD stages and Normal state of kidney with an increment of 4% in accuracy.

The limitation of this work are low classification accuracy and use of smaller number of CKD images. The future work includes, use of greater number of real-world CKD images without using any data augmentation scheme, especially for CKD Stage-4 and CKD Stage-5. Future step also focuses on integration of CKD XAI with LIME, LRP, ATTENTION, SHAP or any other XAI method for the development of a mobile application.

**Acknowledgment:**

Authors are highly grateful to Dr. Kiran Patil (M.D. Radiology) and his team of Radiologists at Shivam diagnostic, Jalgaon, Maharashtra, India for issue of US Normal kidney and CKD images and aiding in completion of legal, medical and ethical requirement for image collection. We are also grateful to Dr. Aleem Ansari Medical Officer at District Civil Hospital, Dhule, Maharashtra, India, for his guidance, and opinions in identification, sorting and result interpretation of images.